\documentclass[11pt]{article}

\usepackage{colacl,amsmath,rotate,psfig}

\newcommand{\tuple}[1]{\langle#1\rangle}

\author{Mats Rooth \\ {\bf  Stefan Riezler} \\ {\bf Detlef
  Prescher} \\ {\bf Glenn Carroll} \\ {\bf Franz Beil} \\
  Institut f\"ur Maschinelle Sprachverarbeitung \\
  University of Stuttgart, Germany}

\title{Inducing a Semantically Annotated Lexicon \\
via EM-Based Clustering}

\begin{document}

\bibliographystyle{acl}

\maketitle

\begin{abstract}
  We present a technique for automatic induction of slot
  annotations for subcategorization frames, based on induction of
  hidden classes in the EM framework of statistical estimation.
  The models are empirically evalutated by a general decision test.
  Induction of slot labeling for subcategorization frames is
  accomplished by a further application of EM, and applied
  experimentally on frame observations derived from parsing large
  corpora. We outline an interpretation
  of the learned representations as theoretical-linguistic
  decompositional lexical entries.
\end{abstract}

\section{Introduction}

An important challenge in computational linguistics concerns the
construction of large-scale computational lexicons for the numerous
natural languages where very large samples of language use are now
available. \newcite{Resnik:93} initiated research into the automatic
acquisition of semantic selectional restrictions. \newcite{Ribas:94}
presented an approach which takes into account the syntactic position
of the elements whose semantic relation is to be acquired. However,
those and most of the following approaches require as
a prerequisite a fixed taxonomy of semantic relations.
This is a problem because (i) entailment hierarchies are presently
available for few languages, and (ii) we regard it as an open
question whether and to what degree existing designs for lexical
hierarchies are appropriate for representing lexical meaning. Both of
these considerations suggest the relevance of inductive and
experimental approaches to the construction of lexicons with semantic
information.

This paper presents a method for automatic induction of semantically
annotated subcategorization frames from unannotated corpora.
We use a statistical subcat-induction system which estimates probability
distributions and corpus frequencies for pairs of a head and a
subcat frame \cite{CarrollRooth:98}.
The statistical parser can also collect frequencies for
the nominal fillers of slots in a subcat frame.  The induction of labels
for slots in a frame is based upon estimation of a probability
distribution over tuples consisting of a class label, a selecting
head, a grammatical relation, and a filler head. The class label is
treated as hidden data in the EM-framework for statistical estimation.

\begin{figure*}[ht]
\begin{center}
\setlength{\tabcolsep}{2pt}

{\tiny
\begin{tabular}
{|l|r|rrrrrrrrrrrrrrrrrrrrrrrrrrrrrr|} \hline
\begin{tabular}{c}
     {\bf Class 17} \\
                                 \\
     PROB 0.0265\\
                                 \\
\end{tabular}
&  & \rotate[l]{0.0379} 
 & \rotate[l]{0.0315} 
 & \rotate[l]{0.0313} 
 & \rotate[l]{0.0249} 
 & \rotate[l]{0.0164} 
 & \rotate[l]{0.0143} 
 & \rotate[l]{0.0110} 
 & \rotate[l]{0.0109} 
 & \rotate[l]{0.0105} 
 & \rotate[l]{0.0103} 
 & \rotate[l]{0.0099} 
 & \rotate[l]{0.0091} 
 & \rotate[l]{0.0089} 
 & \rotate[l]{0.0088} 
 & \rotate[l]{0.0082} 
 & \rotate[l]{0.0077} 
 & \rotate[l]{0.0073} 
 & \rotate[l]{0.0071} 
 & \rotate[l]{0.0071} 
 & \rotate[l]{0.0070} 
 & \rotate[l]{0.0068} 
 & \rotate[l]{0.0067} 
 & \rotate[l]{0.0065} 
 & \rotate[l]{0.0065} 
 & \rotate[l]{0.0058} 
 & \rotate[l]{0.0057} 
 & \rotate[l]{0.0057} 
 & \rotate[l]{0.0054} 
 & \rotate[l]{0.0051} 
 & \rotate[l]{0.0050} 
 \\ 
 \hline
&  & \rotate[l]{number} 
 & \rotate[l]{rate} 
 & \rotate[l]{price} 
 & \rotate[l]{cost} 
 & \rotate[l]{level} 
 & \rotate[l]{amount} 
 & \rotate[l]{sale} 
 & \rotate[l]{value} 
 & \rotate[l]{interest} 
 & \rotate[l]{demand} 
 & \rotate[l]{chance} 
 & \rotate[l]{standard} 
 & \rotate[l]{share} 
 & \rotate[l]{risk} 
 & \rotate[l]{profit} 
 & \rotate[l]{pressure} 
 & \rotate[l]{income} 
 & \rotate[l]{performance} 
 & \rotate[l]{benefit} 
 & \rotate[l]{size} 
 & \rotate[l]{population} 
 & \rotate[l]{proportion} 
 & \rotate[l]{temperature} 
 & \rotate[l]{tax} 
 & \rotate[l]{fee} 
 & \rotate[l]{time} 
 & \rotate[l]{power} 
 & \rotate[l]{quality} 
 & \rotate[l]{supplely} 
 & \rotate[l]{money} 
 \\ 
 \hline
0.0437 & increase.as:s & $\bullet$ & $\bullet$ & $\bullet$ & $\bullet$ & $\bullet$ & $\bullet$ & $\bullet$ & $\bullet$ & $\bullet$ & $\bullet$ & $\bullet$ & $\bullet$ & $\bullet$ & $\bullet$ & $\bullet$ & $\bullet$ & $\bullet$ & $\bullet$ & $\bullet$ & $\bullet$ & $\bullet$ & $\bullet$ & $\bullet$ & $\bullet$ & $\bullet$ & $\bullet$ & $\bullet$ & $\bullet$ & $\bullet$ & \\
0.0392 & increase.aso:o &  & $\bullet$ & $\bullet$ & $\bullet$ & $\bullet$ & $\bullet$ & $\bullet$ & $\bullet$ & $\bullet$ & $\bullet$ & $\bullet$ & $\bullet$ & $\bullet$ & $\bullet$ & $\bullet$ & $\bullet$ & $\bullet$ & $\bullet$ & $\bullet$ & $\bullet$ & $\bullet$ & $\bullet$ &  & $\bullet$ &  & $\bullet$ & $\bullet$ & $\bullet$ & $\bullet$ & $\bullet$\\
0.0344 & fall.as:s & $\bullet$ & $\bullet$ & $\bullet$ & $\bullet$ & $\bullet$ & $\bullet$ & $\bullet$ & $\bullet$ &  & $\bullet$ & $\bullet$ & $\bullet$ & $\bullet$ &  & $\bullet$ & $\bullet$ & $\bullet$ & $\bullet$ &  &  & $\bullet$ & $\bullet$ & $\bullet$ & $\bullet$ &  & $\bullet$ &  & $\bullet$ & $\bullet$ & \\
0.0337 & pay.aso:o & $\bullet$ & $\bullet$ & $\bullet$ &  & $\bullet$ & $\bullet$ &  & $\bullet$ & $\bullet$ & $\bullet$ &  &  & $\bullet$ &  &  &  & $\bullet$ & $\bullet$ & $\bullet$ &  &  & $\bullet$ &  &  & $\bullet$ & $\bullet$ &  &  &  & $\bullet$\\
0.0329 & reduce.aso:o & $\bullet$ & $\bullet$ & $\bullet$ & $\bullet$ & $\bullet$ & $\bullet$ & $\bullet$ & $\bullet$ & $\bullet$ & $\bullet$ & $\bullet$ & $\bullet$ & $\bullet$ & $\bullet$ & $\bullet$ & $\bullet$ & $\bullet$ & $\bullet$ & $\bullet$ & $\bullet$ & $\bullet$ & $\bullet$ & $\bullet$ & $\bullet$ & $\bullet$ & $\bullet$ & $\bullet$ & $\bullet$ & $\bullet$ & \\
0.0257 & rise.as:s & $\bullet$ & $\bullet$ & $\bullet$ & $\bullet$ & $\bullet$ & $\bullet$ & $\bullet$ & $\bullet$ &  & $\bullet$ &  & $\bullet$ & $\bullet$ & $\bullet$ & $\bullet$ & $\bullet$ & $\bullet$ &  & $\bullet$ & $\bullet$ & $\bullet$ & $\bullet$ & $\bullet$ & $\bullet$ &  & $\bullet$ & $\bullet$ & $\bullet$ & $\bullet$ & \\
0.0196 & exceed.aso:o & $\bullet$ & $\bullet$ & $\bullet$ & $\bullet$ & $\bullet$ & $\bullet$ &  & $\bullet$ & $\bullet$ & $\bullet$ & $\bullet$ & $\bullet$ &  &  & $\bullet$ &  & $\bullet$ &  & $\bullet$ & $\bullet$ & $\bullet$ & $\bullet$ &  & $\bullet$ & $\bullet$ & $\bullet$ & $\bullet$ & $\bullet$ & $\bullet$ & \\
0.0177 & exceed.aso:s & $\bullet$ & $\bullet$ & $\bullet$ & $\bullet$ & $\bullet$ & $\bullet$ & $\bullet$ & $\bullet$ & $\bullet$ & $\bullet$ &  &  & $\bullet$ & $\bullet$ & $\bullet$ & $\bullet$ & $\bullet$ & $\bullet$ & $\bullet$ & $\bullet$ & $\bullet$ & $\bullet$ & $\bullet$ & $\bullet$ &  & $\bullet$ &  &  & $\bullet$ & $\bullet$\\
0.0169 & affect.aso:o & $\bullet$ & $\bullet$ & $\bullet$ & $\bullet$ & $\bullet$ & $\bullet$ & $\bullet$ & $\bullet$ & $\bullet$ & $\bullet$ & $\bullet$ & $\bullet$ & $\bullet$ & $\bullet$ & $\bullet$ & $\bullet$ & $\bullet$ & $\bullet$ & $\bullet$ &  & $\bullet$ & $\bullet$ & $\bullet$ &  & $\bullet$ & $\bullet$ & $\bullet$ & $\bullet$ & $\bullet$ & \\
0.0156 & grow.as:s & $\bullet$ &  &  &  &  &  & $\bullet$ & $\bullet$ & $\bullet$ & $\bullet$ &  &  & $\bullet$ & $\bullet$ & $\bullet$ & $\bullet$ & $\bullet$ &  & $\bullet$ & $\bullet$ & $\bullet$ & $\bullet$ &  &  &  &  & $\bullet$ &  &  & $\bullet$\\
0.0134 & include.aso:s & $\bullet$ & $\bullet$ & $\bullet$ & $\bullet$ & $\bullet$ & $\bullet$ & $\bullet$ & $\bullet$ & $\bullet$ & $\bullet$ &  & $\bullet$ & $\bullet$ & $\bullet$ & $\bullet$ & $\bullet$ & $\bullet$ & $\bullet$ & $\bullet$ &  & $\bullet$ & $\bullet$ &  &  & $\bullet$ & $\bullet$ & $\bullet$ & $\bullet$ & $\bullet$ & $\bullet$\\
0.0129 & reach.aso:s & $\bullet$ & $\bullet$ & $\bullet$ & $\bullet$ & $\bullet$ & $\bullet$ & $\bullet$ &  & $\bullet$ & $\bullet$ &  &  & $\bullet$ &  & $\bullet$ & $\bullet$ &  & $\bullet$ & $\bullet$ &  & $\bullet$ & $\bullet$ & $\bullet$ &  &  & $\bullet$ & $\bullet$ & $\bullet$ & $\bullet$ & $\bullet$\\
0.0120 & decline.as:s & $\bullet$ & $\bullet$ & $\bullet$ & $\bullet$ & $\bullet$ &  & $\bullet$ & $\bullet$ & $\bullet$ & $\bullet$ &  & $\bullet$ & $\bullet$ & $\bullet$ & $\bullet$ &  &  &  &  & $\bullet$ & $\bullet$ & $\bullet$ & $\bullet$ &  &  &  &  & $\bullet$ & $\bullet$ & \\
0.0102 & lose.aso:o & $\bullet$ & $\bullet$ &  & $\bullet$ &  & $\bullet$ & $\bullet$ & $\bullet$ & $\bullet$ &  & $\bullet$ & $\bullet$ & $\bullet$ & $\bullet$ & $\bullet$ & $\bullet$ & $\bullet$ &  & $\bullet$ &  & $\bullet$ & $\bullet$ & $\bullet$ &  & $\bullet$ & $\bullet$ & $\bullet$ & $\bullet$ &  & $\bullet$\\
0.0099 & act.aso:s &  & $\bullet$ & $\bullet$ &  &  &  & $\bullet$ &  &  &  &  & $\bullet$ &  &  &  &  &  &  & $\bullet$ &  &  &  &  & $\bullet$ &  &  & $\bullet$ &  & $\bullet$ & \\
0.0099 & improve.aso:o & $\bullet$ & $\bullet$ &  &  & $\bullet$ &  &  & $\bullet$ &  &  & $\bullet$ & $\bullet$ & $\bullet$ &  & $\bullet$ &  &  & $\bullet$ &  &  &  &  &  &  &  & $\bullet$ & $\bullet$ & $\bullet$ &  & \\
0.0088 & include.aso:o & $\bullet$ & $\bullet$ & $\bullet$ & $\bullet$ & $\bullet$ & $\bullet$ & $\bullet$ & $\bullet$ & $\bullet$ & $\bullet$ &  & $\bullet$ & $\bullet$ & $\bullet$ & $\bullet$ & $\bullet$ & $\bullet$ & $\bullet$ & $\bullet$ & $\bullet$ & $\bullet$ & $\bullet$ & $\bullet$ & $\bullet$ & $\bullet$ & $\bullet$ & $\bullet$ & $\bullet$ & $\bullet$ & $\bullet$\\
0.0088 & cut.aso:o & $\bullet$ & $\bullet$ & $\bullet$ & $\bullet$ & $\bullet$ &  & $\bullet$ & $\bullet$ &  &  & $\bullet$ & $\bullet$ &  & $\bullet$ & $\bullet$ &  & $\bullet$ &  & $\bullet$ &  &  &  &  & $\bullet$ & $\bullet$ & $\bullet$ & $\bullet$ &  &  & \\
0.0080 & show.aso:o & $\bullet$ & $\bullet$ &  & $\bullet$ & $\bullet$ & $\bullet$ & $\bullet$ & $\bullet$ & $\bullet$ & $\bullet$ &  & $\bullet$ & $\bullet$ & $\bullet$ & $\bullet$ & $\bullet$ & $\bullet$ & $\bullet$ & $\bullet$ & $\bullet$ & $\bullet$ & $\bullet$ &  &  &  & $\bullet$ & $\bullet$ & $\bullet$ &  & \\
0.0078 & vary.as:s & $\bullet$ & $\bullet$ & $\bullet$ & $\bullet$ & $\bullet$ & $\bullet$ &  & $\bullet$ & $\bullet$ & $\bullet$ &  & $\bullet$ &  &  &  &  &  & $\bullet$ &  & $\bullet$ &  & $\bullet$ & $\bullet$ &  &  & $\bullet$ &  & $\bullet$ &  & \\
\hline\end{tabular}}

%%% Local Variables: 
%%% mode: latex
%%% TeX-master: t
%%% TeX-master: "/projekte/gramotron/docSrc/acl99LSC/paper-final"
%%% End: 

 \end{center}
  \caption{Class 17: scalar change}
  \label{class17}
\end{figure*}

\section{EM-Based Clustering}
\label{s:cluster}

In our clustering approach, classes are derived directly from
distributional data---a sample of pairs of verbs and nouns, gathered
by parsing an unannotated corpus and extracting the fillers of
grammatical relations. Semantic classes corresponding to such pairs
are viewed as hidden variables or unobserved data in the context of
maximum likelihood estimation from incomplete data via the EM
algorithm. This approach allows us to work in a mathematically
well-defined framework of statistical inference, i.e., standard
monotonicity and convergence results for the EM algorithm extend to
our method. The two main tasks of EM-based clustering are i) the
induction of a smooth probability model on the data, and ii) the
automatic discovery of class-structure in the data. Both of these
aspects are respected in our application of lexicon induction.
The basic ideas of our EM-based
clustering approach were presented in \newcite{Rooth:95}. 
Our approach contrasts with the merely
heuristic and empirical justification of similarity-based approaches
to clustering \cite{Dagan:98} for which so far no clear probabilistic
interpretation has been given. The
probability model we use can be found earlier in \newcite{Pereira:93}.
However, in contrast to this approach, our statistical inference
method for clustering is formalized clearly as an EM-algorithm.
Approaches to probabilistic clustering similar to ours were presented
recently in \newcite{Saul:97} and \newcite{Hofmann:98}. There also
EM-algorithms for similar probability models have been derived, but
applied only to simpler tasks not involving a combination of EM-based
clustering models as in our lexicon induction experiment. For further
applications of our clustering model see \newcite{Rooth:98}.

\begin{figure*}[ht]
\begin{center}
\setlength{\tabcolsep}{2pt}

{\tiny
\begin{tabular}
{|l|r|rrrrrrrrrrrrrrrrrrrrrrrrrrrrrr|} \hline
\begin{tabular}{c}
     {\bf Class 5} \\
                                 \\
     PROB 0.0412\\
                                 \\
\end{tabular}
&  & \rotate[l]{0.0148} 
 & \rotate[l]{0.0084} 
 & \rotate[l]{0.0082} 
 & \rotate[l]{0.0078} 
 & \rotate[l]{0.0074} 
 & \rotate[l]{0.0071} 
 & \rotate[l]{0.0054} 
 & \rotate[l]{0.0049} 
 & \rotate[l]{0.0048} 
 & \rotate[l]{0.0047} 
 & \rotate[l]{0.0046} 
 & \rotate[l]{0.0041} 
 & \rotate[l]{0.0040} 
 & \rotate[l]{0.0040} 
 & \rotate[l]{0.0039} 
 & \rotate[l]{0.0039} 
 & \rotate[l]{0.0039} 
 & \rotate[l]{0.0039} 
 & \rotate[l]{0.0038} 
 & \rotate[l]{0.0038} 
 & \rotate[l]{0.0038} 
 & \rotate[l]{0.0037} 
 & \rotate[l]{0.0035} 
 & \rotate[l]{0.0035} 
 & \rotate[l]{0.0035} 
 & \rotate[l]{0.0034} 
 & \rotate[l]{0.0033} 
 & \rotate[l]{0.0033} 
 & \rotate[l]{0.0033} 
 & \rotate[l]{0.0033} 
 \\ 
 \hline
&  & \rotate[l]{man} 
 & \rotate[l]{ruth} 
 & \rotate[l]{corbett} 
 & \rotate[l]{doctor} 
 & \rotate[l]{woman} 
 & \rotate[l]{athelstan} 
 & \rotate[l]{cranston} 
 & \rotate[l]{benjamin} 
 & \rotate[l]{stephen} 
 & \rotate[l]{adam} 
 & \rotate[l]{girl} 
 & \rotate[l]{laura} 
 & \rotate[l]{maggie} 
 & \rotate[l]{voice} 
 & \rotate[l]{john} 
 & \rotate[l]{harry} 
 & \rotate[l]{emily} 
 & \rotate[l]{one} 
 & \rotate[l]{people} 
 & \rotate[l]{boy} 
 & \rotate[l]{rachel} 
 & \rotate[l]{ashley} 
 & \rotate[l]{jane} 
 & \rotate[l]{caroline} 
 & \rotate[l]{jack} 
 & \rotate[l]{burun} 
 & \rotate[l]{juliet} 
 & \rotate[l]{blanche} 
 & \rotate[l]{helen} 
 & \rotate[l]{edward} 
 \\ 
 \hline
0.0542 & ask.as:s & $\bullet$ & $\bullet$ & $\bullet$ & $\bullet$ & $\bullet$ & $\bullet$ & $\bullet$ & $\bullet$ & $\bullet$ & $\bullet$ & $\bullet$ & $\bullet$ & $\bullet$ & $\bullet$ & $\bullet$ & $\bullet$ & $\bullet$ & $\bullet$ & $\bullet$ & $\bullet$ & $\bullet$ & $\bullet$ & $\bullet$ &  & $\bullet$ & $\bullet$ & $\bullet$ & $\bullet$ & $\bullet$ & $\bullet$\\
0.0340 & nod.as:s &  & $\bullet$ & $\bullet$ & $\bullet$ & $\bullet$ & $\bullet$ & $\bullet$ & $\bullet$ & $\bullet$ &  & $\bullet$ & $\bullet$ & $\bullet$ &  & $\bullet$ & $\bullet$ & $\bullet$ &  &  & $\bullet$ & $\bullet$ & $\bullet$ & $\bullet$ & $\bullet$ & $\bullet$ & $\bullet$ & $\bullet$ & $\bullet$ & $\bullet$ & $\bullet$\\
0.0299 & think.as:s & $\bullet$ & $\bullet$ & $\bullet$ & $\bullet$ & $\bullet$ & $\bullet$ &  &  & $\bullet$ & $\bullet$ & $\bullet$ & $\bullet$ & $\bullet$ &  & $\bullet$ & $\bullet$ & $\bullet$ & $\bullet$ & $\bullet$ & $\bullet$ & $\bullet$ & $\bullet$ & $\bullet$ & $\bullet$ & $\bullet$ &  & $\bullet$ & $\bullet$ & $\bullet$ & $\bullet$\\
0.0287 & shake.aso:s & $\bullet$ & $\bullet$ & $\bullet$ & $\bullet$ & $\bullet$ & $\bullet$ & $\bullet$ & $\bullet$ & $\bullet$ & $\bullet$ & $\bullet$ & $\bullet$ & $\bullet$ &  & $\bullet$ & $\bullet$ & $\bullet$ &  & $\bullet$ & $\bullet$ & $\bullet$ & $\bullet$ & $\bullet$ & $\bullet$ & $\bullet$ & $\bullet$ & $\bullet$ &  & $\bullet$ & $\bullet$\\
0.0264 & smile.as:s & $\bullet$ & $\bullet$ & $\bullet$ & $\bullet$ & $\bullet$ & $\bullet$ & $\bullet$ & $\bullet$ & $\bullet$ & $\bullet$ & $\bullet$ & $\bullet$ & $\bullet$ &  & $\bullet$ & $\bullet$ & $\bullet$ & $\bullet$ & $\bullet$ & $\bullet$ & $\bullet$ & $\bullet$ & $\bullet$ & $\bullet$ &  & $\bullet$ & $\bullet$ & $\bullet$ & $\bullet$ & $\bullet$\\
0.0213 & laugh.as:s & $\bullet$ & $\bullet$ &  & $\bullet$ &  & $\bullet$ & $\bullet$ & $\bullet$ & $\bullet$ & $\bullet$ & $\bullet$ & $\bullet$ & $\bullet$ & $\bullet$ & $\bullet$ & $\bullet$ & $\bullet$ & $\bullet$ & $\bullet$ & $\bullet$ & $\bullet$ & $\bullet$ & $\bullet$ & $\bullet$ & $\bullet$ & $\bullet$ & $\bullet$ & $\bullet$ & $\bullet$ & $\bullet$\\
0.0207 & reply.as:s & $\bullet$ & $\bullet$ & $\bullet$ & $\bullet$ & $\bullet$ & $\bullet$ & $\bullet$ & $\bullet$ & $\bullet$ & $\bullet$ & $\bullet$ & $\bullet$ & $\bullet$ & $\bullet$ & $\bullet$ & $\bullet$ & $\bullet$ & $\bullet$ & $\bullet$ & $\bullet$ & $\bullet$ & $\bullet$ &  &  & $\bullet$ &  &  &  & $\bullet$ & $\bullet$\\
0.0167 & shrug.as:s & $\bullet$ &  & $\bullet$ & $\bullet$ & $\bullet$ & $\bullet$ & $\bullet$ & $\bullet$ & $\bullet$ & $\bullet$ & $\bullet$ & $\bullet$ & $\bullet$ &  &  & $\bullet$ & $\bullet$ &  &  & $\bullet$ & $\bullet$ &  & $\bullet$ & $\bullet$ & $\bullet$ & $\bullet$ & $\bullet$ & $\bullet$ & $\bullet$ & $\bullet$\\
0.0148 & wonder.as:s & $\bullet$ & $\bullet$ & $\bullet$ & $\bullet$ & $\bullet$ & $\bullet$ &  &  & $\bullet$ & $\bullet$ & $\bullet$ &  &  &  & $\bullet$ & $\bullet$ & $\bullet$ &  & $\bullet$ & $\bullet$ & $\bullet$ & $\bullet$ & $\bullet$ & $\bullet$ & $\bullet$ & $\bullet$ & $\bullet$ & $\bullet$ & $\bullet$ & $\bullet$\\
0.0141 & feel.aso:s & $\bullet$ &  & $\bullet$ & $\bullet$ & $\bullet$ &  &  &  & $\bullet$ & $\bullet$ & $\bullet$ & $\bullet$ & $\bullet$ &  & $\bullet$ & $\bullet$ & $\bullet$ & $\bullet$ & $\bullet$ & $\bullet$ & $\bullet$ & $\bullet$ & $\bullet$ & $\bullet$ & $\bullet$ &  & $\bullet$ & $\bullet$ & $\bullet$ & $\bullet$\\
0.0133 & take.aso:s & $\bullet$ & $\bullet$ & $\bullet$ & $\bullet$ & $\bullet$ &  & $\bullet$ & $\bullet$ & $\bullet$ & $\bullet$ & $\bullet$ & $\bullet$ & $\bullet$ & $\bullet$ & $\bullet$ & $\bullet$ & $\bullet$ & $\bullet$ & $\bullet$ & $\bullet$ & $\bullet$ & $\bullet$ & $\bullet$ & $\bullet$ & $\bullet$ &  & $\bullet$ & $\bullet$ & $\bullet$ & $\bullet$\\
0.0121 & sigh.as:s & $\bullet$ & $\bullet$ & $\bullet$ & $\bullet$ & $\bullet$ & $\bullet$ &  & $\bullet$ & $\bullet$ & $\bullet$ &  & $\bullet$ &  &  & $\bullet$ & $\bullet$ & $\bullet$ &  & $\bullet$ &  & $\bullet$ & $\bullet$ & $\bullet$ & $\bullet$ & $\bullet$ & $\bullet$ &  & $\bullet$ & $\bullet$ & $\bullet$\\
0.0110 & watch.aso:s & $\bullet$ & $\bullet$ & $\bullet$ &  & $\bullet$ & $\bullet$ & $\bullet$ & $\bullet$ & $\bullet$ & $\bullet$ & $\bullet$ &  & $\bullet$ &  & $\bullet$ & $\bullet$ & $\bullet$ & $\bullet$ & $\bullet$ & $\bullet$ & $\bullet$ &  & $\bullet$ &  & $\bullet$ & $\bullet$ & $\bullet$ &  & $\bullet$ & \\
0.0106 & ask.aso:s & $\bullet$ & $\bullet$ & $\bullet$ & $\bullet$ & $\bullet$ & $\bullet$ & $\bullet$ & $\bullet$ & $\bullet$ & $\bullet$ & $\bullet$ & $\bullet$ & $\bullet$ & $\bullet$ & $\bullet$ & $\bullet$ & $\bullet$ & $\bullet$ &  & $\bullet$ &  &  & $\bullet$ &  & $\bullet$ &  &  & $\bullet$ & $\bullet$ & $\bullet$\\
0.0104 & tell.aso:s & $\bullet$ & $\bullet$ & $\bullet$ & $\bullet$ & $\bullet$ &  & $\bullet$ & $\bullet$ & $\bullet$ & $\bullet$ & $\bullet$ & $\bullet$ & $\bullet$ & $\bullet$ & $\bullet$ & $\bullet$ &  & $\bullet$ & $\bullet$ & $\bullet$ & $\bullet$ & $\bullet$ & $\bullet$ & $\bullet$ & $\bullet$ & $\bullet$ & $\bullet$ & $\bullet$ & $\bullet$ & $\bullet$\\
0.0094 & look.as:s & $\bullet$ & $\bullet$ & $\bullet$ & $\bullet$ & $\bullet$ & $\bullet$ & $\bullet$ & $\bullet$ & $\bullet$ & $\bullet$ & $\bullet$ &  &  &  & $\bullet$ &  & $\bullet$ & $\bullet$ & $\bullet$ & $\bullet$ & $\bullet$ &  &  &  & $\bullet$ & $\bullet$ & $\bullet$ &  &  & $\bullet$\\
0.0092 & give.aso:s & $\bullet$ & $\bullet$ &  & $\bullet$ & $\bullet$ &  &  & $\bullet$ & $\bullet$ & $\bullet$ & $\bullet$ & $\bullet$ & $\bullet$ & $\bullet$ & $\bullet$ & $\bullet$ & $\bullet$ & $\bullet$ & $\bullet$ & $\bullet$ & $\bullet$ & $\bullet$ & $\bullet$ & $\bullet$ & $\bullet$ & $\bullet$ & $\bullet$ & $\bullet$ & $\bullet$ & $\bullet$\\
0.0089 & hear.aso:s & $\bullet$ & $\bullet$ & $\bullet$ & $\bullet$ & $\bullet$ & $\bullet$ & $\bullet$ &  &  &  & $\bullet$ &  & $\bullet$ & $\bullet$ & $\bullet$ & $\bullet$ &  & $\bullet$ & $\bullet$ & $\bullet$ &  &  & $\bullet$ & $\bullet$ & $\bullet$ & $\bullet$ & $\bullet$ & $\bullet$ &  & \\
0.0083 & grin.as:s & $\bullet$ & $\bullet$ & $\bullet$ & $\bullet$ & $\bullet$ & $\bullet$ & $\bullet$ & $\bullet$ &  & $\bullet$ & $\bullet$ & $\bullet$ & $\bullet$ &  & $\bullet$ & $\bullet$ & $\bullet$ &  &  & $\bullet$ & $\bullet$ & $\bullet$ & $\bullet$ & $\bullet$ & $\bullet$ & $\bullet$ &  &  & $\bullet$ & $\bullet$\\
0.0083 & answer.as:s & $\bullet$ & $\bullet$ & $\bullet$ & $\bullet$ & $\bullet$ & $\bullet$ & $\bullet$ & $\bullet$ &  &  &  &  & $\bullet$ & $\bullet$ & $\bullet$ & $\bullet$ & $\bullet$ & $\bullet$ & $\bullet$ & $\bullet$ &  & $\bullet$ &  & $\bullet$ & $\bullet$ &  & $\bullet$ &  & $\bullet$ & $\bullet$\\
\hline\end{tabular}}

%%% Local Variables: 
%%% mode: latex
%%% TeX-master: t
%%% TeX-master: "/projekte/gramotron/docSrc/acl99LSC/paper-final"
%%% TeX-master: "/projekte/gramotron/docSrc/acl99LSC/paper-final"
%%% End: 

 \end{center}
  \caption{Class 5: communicative action}
  \label{class5}
\end{figure*}

We seek to derive a joint distribution of verb-noun pairs from a large
sample of pairs of verbs $v \in V$ and nouns $n \in N$. The key idea
is to view $v$ and $n$ as conditioned on a hidden class $c \in C$,
where the classes are given no prior interpretation.
The semantically smoothed probability of a pair $(v,n)$ is defined to be:
\[
p(v,n) = \sum_{c \in C} p(c,v,n) = \sum_{c \in C} p(c) p(v|c) p(n|c)
\]
The joint distribution $p(c,v,n)$ is defined by
\(
p(c,v,n) = p(c) p(v|c) p(n|c)
\).
Note that by construction, conditioning of
$v$ and $n$ on each other is solely made through the classes $c$.

In the framework of the EM algorithm \cite{Dempster:77},
we can formalize clustering as an estimation problem for a latent
class (LC) model as follows. We
are given:
(i) a sample space $\mathcal{Y}$ of observed, incomplete data,
  corresponding to pairs from $V \times N$,
(ii) a sample space $\mathcal{X}$ of unobserved, complete data,
  corresponding to triples from $C \times V \times N$,
(iii) a set $X(y) = \{ x \in \mathcal{X} \enspace | \enspace x = (c,y),  \; c \in C \}$ of complete data related to the observation $y$,
(iv) a complete-data specification $p_\theta(x)$, corresponding to the joint probability
  $p(c,v,n)$ over $C \times V \times N$, with parameter-vector $\theta
  = \tuple{\theta_c, \theta_{vc}, \theta_{nc} | c \in C, v \in V, n \in N }$,
(v) an incomplete data specification $p_\theta(y)$ which is related
  to the complete-data specification as the marginal probability
$ p_\theta(y) = \sum_{X(y)} p_\theta(x).$

The EM algorithm is directed at finding a value $\hat\theta$ of
$\theta$ that maximizes the incomplete-data log-likelihood
function $L$ as a function of $\theta$ for a given sample $\mathcal{Y}$, i.e., 
\( 
\hat\theta = \underset{\theta}{\arg\max\;} L(\theta) 
\textrm{ where } L(\theta) =  \ln \prod_{y}
p_\theta(y).
\) 

As prescribed by the EM algorithm, the parameters of $L(\theta)$ are
estimated indirectly by proceeding iteratively in terms of
complete-data estimation for the auxiliary function $Q(\theta; \theta^{(t)})$,
which is the conditional expectation of the complete-data
log-likelihood $\ln p_\theta(x)$ given the observed data $y$ and the
current fit of the parameter values $\theta^{(t)}$ (E-step). This
auxiliary function is iteratively maximized as a function of $\theta$
(M-step), where each iteration is defined by the map
\(
\theta^{(t+1)} = M(\theta^{(t)}) = \underset{\theta}{\arg\max\;}Q(\theta;\theta^{(t)})
\)
Note that our application is an instance of the EM-algorithm for
context-free models \cite{Baum:70}, from which the following
particularly simple reestimation formulae can be derived. Let
$x=(c,y)$, and $f(y)$ the sample-frequency of $y$. Then
\begin{eqnarray*}
M(\theta_{vc}) & = & \frac{\sum_{y \in \{v\} \times N} f(y) p_\theta(x|y)}
{\sum_{y}f(y) p_\theta(x|y)}, \\ 
M(\theta_{nc}) & = & \frac{\sum_{y \in V \times \{n\}}f(y) p_\theta(x|y)}
{\sum_{y}f(y) p_\theta(x|y)}, \\ 
M(\theta_{c}) & = & \frac{\sum_{y}f(y) p_\theta(x|y)}
{|\mathcal{Y}|}.
\end{eqnarray*}
Intuitively, the conditional expectation of the number of times a particular $v$, $n$, or
$c$ choice is made during the derivation is prorated by the
conditionally expected total number of times a choice of the same kind
is made. As shown by \newcite{Baum:70}, these expectations can be
calculated efficiently using dynamic programming techniques. Every
such maximization step increases the log-likelihood function $L$, and
a sequence of re-estimates eventually converges to a (local) maximum of $L$.

In the following, we will present some examples of induced clusters. 
Input to the clustering algorithm was a training corpus of 1280715
tokens (608850 types) of verb-noun pairs participating in the
grammatical relations of intransitive and transitive verbs and their
subject- and object-fillers. The data were gathered from the
maximal-probability parses the head-lexicalized probabilistic
context-free grammar of \cite{CarrollRooth:98} gave for the British
National Corpus (117 million words).

\begin{figure}[htbp]
\begin{center}
\mbox{\psfig{file=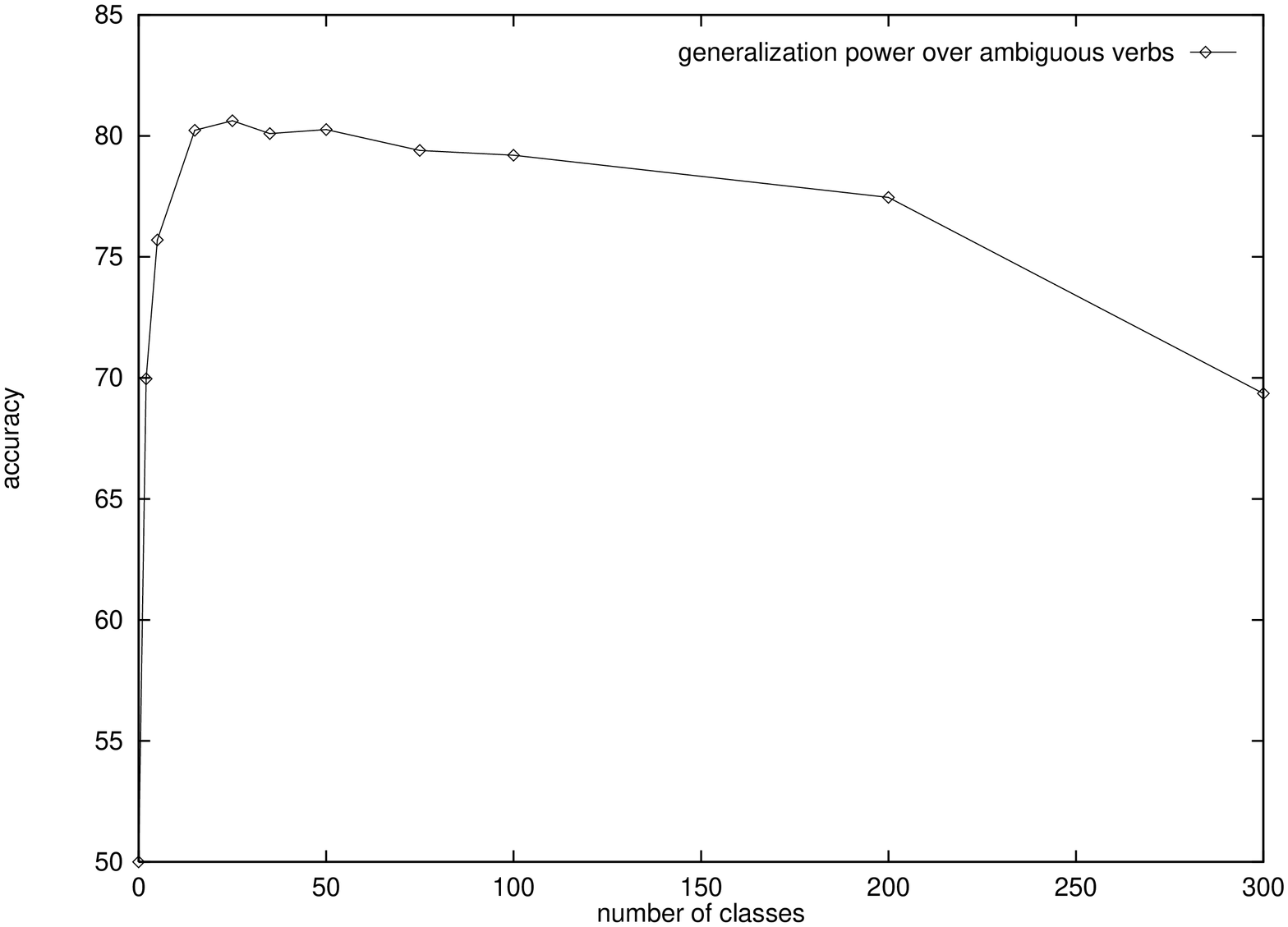,angle=270,width=7cm,height=3.5cm}}
\end{center}
\caption{Evaluation of pseudo-disambiguation}
\label{eval}
\end{figure}

Fig. \ref{class5} shows an induced semantic class out of a model with
35 classes. At the top are listed the 20 most probable nouns in the
$p(n|5)$ distribution and their probabilities, and at left are the 30
most probable verbs in the $p(v|5)$ distribution. 5 is the class
index. Those verb-noun pairs which were seen in the training data
appear with a dot in the class matrix.  Verbs with suffix $.as:s$
indicate the subject slot of an active intransitive. Similarily
$.aso:s$ denotes the subject slot of an active transitive, and
$.aso:o$ denotes the object slot of an active transitive.  Thus $v$
in the above discussion actually consists of a combination of a verb
with a subcat frame slot $as:s$, $aso:s$, or $aso:o$.
Induced classes often have a basis in lexical semantics; class 5 can
be interpreted as clustering agents, denoted by proper names, ``man'',
and ``woman'', together with verbs denoting \emph{communicative action}.
Fig. \ref{class17} shows a cluster involving verbs of \emph{scalar
  change} and things which can move along scales. Fig. \ref{class8}
can be interpreted as involving different \emph{dispositions} and
modes of their execution.

\section{Evaluation of Clustering Models}
\label{s:eval}

\subsection{Pseudo-Disambiguation}

We evaluated our clustering models on a pseudo-disambiguation task similar to that
performed in \newcite{Pereira:93}, but differing in detail.
The task is to judge which of two verbs $v$ and $v'$ is more likely to take a given
noun $n$ as its argument where the pair $(v,n)$ has been cut out of
the original corpus and the pair $(v',n)$ is constructed by
pairing $n$ with a randomly chosen verb $v'$ such that the combination
$(v',n)$ is completely unseen. Thus this test evaluates how well the
models generalize over unseen verbs.

The data for this test were built as follows. We constructed
an evaluation corpus of $(v, n, v')$ triples by randomly cutting a
test corpus of 3000 $(v,n)$ pairs out of the original corpus of 1280712
tokens, leaving a training corpus of 1178698 tokens. Each noun $n$ in the test corpus
was combined with a verb $v'$ which was randomly chosen according to
its frequency such that the pair $(v',n)$ did appear neither in
the training nor in the test corpus. However, the elements $v$, $v'$,
and $n$ were required to be part of the training corpus.
Furthermore, we restricted the verbs and nouns in the evaluation
corpus to the ones which occurred at least
30 times and at most 3000 times with some verb-functor $v$ in the
training corpus. The resulting 1337 evaluation triples were used to
evaluate a sequence of clustering models trained from the training corpus. 

The clustering models we evaluated were parameterized in starting
values of the training algorithm, in the number of classes of the
model, and in the number of iteration steps,
resulting in a sequence of $3 \times 10 \times 6$ models.
Starting from a lower bound of 50 \% random choice, 
accuracy was calculated as the number of times the model decided for
$p(n|v) \geq p(n|v')$ out of all choices made. 
Fig. \ref{eval} shows the evaluation results 
for models trained with 50 iterations, averaged over starting values,
and plotted against class cardinality. Different starting values had
an effect of ${\scriptscriptstyle \overset{+}{-}}$ 2 \% on the performance of
the test. We obtained a value of about 80
\% accuracy for models between 25 and 100 classes. Models with more than 100
classes show a small but stable overfitting effect.

\subsection{Smoothing Power}

\begin{figure}[tb]
\begin{center}
\mbox{\psfig{file=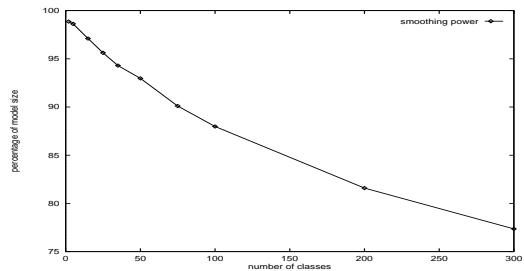,angle=270,width=7cm,height=3.5cm}}
\end{center}
\caption{Evaluation on smoothing task}
\label{smoothing}
\end{figure}

A second experiment addressed the smoothing power of the model by
counting the number of $(v,n)$ pairs in the set $V \times N$ of all possible combinations of verbs and nouns which received a
positive joint probability by the model. The $V \times N$-space
for the above clustering models included about 425 million $(v,n)$
combinations; we approximated the smoothing size of a model by
randomly sampling 1000 pairs from $V \times N$ and returning the
percentage of positively assigned pairs in the random sample. Fig.
\ref{smoothing} plots the smoothing results for the above models against the
number of classes. Starting values had an influence of
${\scriptscriptstyle \overset{+}{-}}$ 1 \% on performance. Given the
proportion of the number of types in the training corpus to the $V
\times N$-space, without clustering we have a smoothing power of 0.14
\% whereas for example a model with 50 classes and 50 iterations has a
smoothing power of about 93 \%. 

Corresponding to the maximum likelihood paradigm,
the number of training iterations had a decreasing effect on the
smoothing performance whereas the accuracy of the
pseudo-disambiguation was increasing in the number of iterations. We
found a number of 50 iterations to be a good compromise in this trade-off.

\section{Lexicon Induction Based on Latent Classes}
\label{s:hsc}

The goal of the following experiment was to derive a lexicon of several hundred
intransitive and transitive verbs with subcat slots 
labeled with latent classes.

\subsection{Probabilistic Labeling with Latent Classes using EM-estimation}

To induce latent classes for the subject slot of a fixed
intransitive verb the following statistical inference step was performed.
Given a latent class model
$p_{LC}(\cdot)$ for verb-noun pairs, and a sample $n_1, \dots, n_M$
of subjects for a fixed intransitive verb, 
we calculate the probability of an arbitrary subject $n \in N$ by:
\[
p(n) = \sum_{c \in C} p(c,n) = \sum_{c \in C} p(c) p_{LC}(n|c).
\]
The estimation of the parameter-vector 
$\theta = \tuple{\theta_c| c \in C}$ can be
formalized in the EM framework by viewing $p(n)$ or $p(c,n)$ as a
function of $\theta$ for fixed $p_{LC}(.)$. The re-estimation
formulae resulting from the incomplete data estimation for these
probability functions have the following form ($f(n)$ is the
frequency of $n$ in the sample of subjects of the fixed verb):
\[
M(\theta_c) = \frac{\sum_{n \in N} f(n) p_\theta(c|n)} {\sum_{n \in N} f(n)}
\]
A similar EM induction process can be applied also to
pairs of nouns, thus enabling induction of latent semantic annotations for
transitive verb frames.
Given a LC model
$p_{LC}(\cdot)$ for verb-noun pairs, and a sample 
$(n_1,n_2)_1, \dots, (n_1,n_2)_M$
of noun arguments ($n_1$ subjects, and $n_2$ direct objects) 
for a fixed transitive verb, 
we calculate the probability of its noun
argument pairs by:
\begin{eqnarray*}
& p(n_1, n_2) = \sum_{c_1,c_2 \in C} p(c_1, c_2, n_1, n_2) \\
 &= \sum_{c_1,c_2 \in C} p(c_1, c_2) p_{LC}(n_1|c_1) p_{LC}(n_2|c_2)
\end{eqnarray*}
Again, estimation of the parameter-vector 
$\theta = \tuple{\theta_{c_1c_2}|c_1, c_2 \in C}$ can be
formalized in an EM framework by viewing $p(n_1,n_2)$ or 
$p(c_1, c_2, n_1, n_2)$ as a
function of $\theta$ for fixed $p_{LC}(.)$. The re-estimation
formulae resulting from this incomplete data estimation problem have
the following simple form ($f(n_1,n_2)$ is the frequency of
$(n_1,n_2)$ in the
sample of noun argument pairs of the fixed verb):
\[
M(\theta_{c1c2}) = \frac{\sum_{n_1,n_2 \in N} f(n_1,n_2)
  p_\theta(c_1,c_2|n_1,n_2)} {\sum_{n_1,n_2 \in N} f(n_1,n_2)}
\]
Note that the class distributions $p(c)$ and $p(c_1, c_2)$ for
intransitive and transitive models can be computed also for verbs
unseen in the LC model.

\begin{figure*}[ht]
\begin{center}
\setlength{\tabcolsep}{2pt}

{\tiny
\begin{tabular}
{|l|r|rrrrrrrrrrrrrrrrrrrrrrrrrrrrrr|} \hline
\begin{tabular}{c}
     {\bf Class 8} \\
                                 \\
     PROB 0.0369\\
                                 \\
\end{tabular}
&  & \rotate[l]{0.0385} 
 & \rotate[l]{0.0162} 
 & \rotate[l]{0.0157} 
 & \rotate[l]{0.0101} 
 & \rotate[l]{0.0073} 
 & \rotate[l]{0.0071} 
 & \rotate[l]{0.0063} 
 & \rotate[l]{0.0060} 
 & \rotate[l]{0.0060} 
 & \rotate[l]{0.0057} 
 & \rotate[l]{0.0055} 
 & \rotate[l]{0.0052} 
 & \rotate[l]{0.0052} 
 & \rotate[l]{0.0051} 
 & \rotate[l]{0.0050} 
 & \rotate[l]{0.0050} 
 & \rotate[l]{0.0048} 
 & \rotate[l]{0.0047} 
 & \rotate[l]{0.0047} 
 & \rotate[l]{0.0046} 
 & \rotate[l]{0.0041} 
 & \rotate[l]{0.0041} 
 & \rotate[l]{0.0040} 
 & \rotate[l]{0.0040} 
 & \rotate[l]{0.0039} 
 & \rotate[l]{0.0038} 
 & \rotate[l]{0.0037} 
 & \rotate[l]{0.0037} 
 & \rotate[l]{0.0036} 
 & \rotate[l]{0.0036} 
 \\ 
 \hline
&  & \rotate[l]{change} 
 & \rotate[l]{use} 
 & \rotate[l]{increase} 
 & \rotate[l]{development} 
 & \rotate[l]{growth} 
 & \rotate[l]{effect} 
 & \rotate[l]{result} 
 & \rotate[l]{degree} 
 & \rotate[l]{response} 
 & \rotate[l]{approach} 
 & \rotate[l]{reduction} 
 & \rotate[l]{forme} 
 & \rotate[l]{condition} 
 & \rotate[l]{understanding} 
 & \rotate[l]{improvement} 
 & \rotate[l]{treatment} 
 & \rotate[l]{skill} 
 & \rotate[l]{action} 
 & \rotate[l]{process} 
 & \rotate[l]{activity} 
 & \rotate[l]{knowledge} 
 & \rotate[l]{factor} 
 & \rotate[l]{level} 
 & \rotate[l]{type} 
 & \rotate[l]{reaction} 
 & \rotate[l]{kind} 
 & \rotate[l]{difference} 
 & \rotate[l]{movement} 
 & \rotate[l]{loss} 
 & \rotate[l]{amount} 
 \\ 
 \hline
0.0539 & require.aso:o & $\bullet$ & $\bullet$ & $\bullet$ & $\bullet$ & $\bullet$ & $\bullet$ & $\bullet$ & $\bullet$ & $\bullet$ & $\bullet$ & $\bullet$ & $\bullet$ & $\bullet$ & $\bullet$ & $\bullet$ & $\bullet$ & $\bullet$ & $\bullet$ & $\bullet$ & $\bullet$ & $\bullet$ & $\bullet$ &  & $\bullet$ &  &  & $\bullet$ & $\bullet$ &  & $\bullet$\\
0.0469 & show.aso:o & $\bullet$ & $\bullet$ & $\bullet$ & $\bullet$ & $\bullet$ & $\bullet$ & $\bullet$ & $\bullet$ & $\bullet$ & $\bullet$ & $\bullet$ & $\bullet$ & $\bullet$ & $\bullet$ & $\bullet$ & $\bullet$ & $\bullet$ & $\bullet$ & $\bullet$ & $\bullet$ & $\bullet$ &  & $\bullet$ & $\bullet$ & $\bullet$ &  & $\bullet$ & $\bullet$ & $\bullet$ & $\bullet$\\
0.0439 & need.aso:o & $\bullet$ & $\bullet$ & $\bullet$ & $\bullet$ &  & $\bullet$ & $\bullet$ & $\bullet$ & $\bullet$ & $\bullet$ & $\bullet$ & $\bullet$ & $\bullet$ & $\bullet$ & $\bullet$ & $\bullet$ & $\bullet$ & $\bullet$ &  & $\bullet$ & $\bullet$ &  & $\bullet$ & $\bullet$ & $\bullet$ & $\bullet$ & $\bullet$ & $\bullet$ &  & $\bullet$\\
0.0383 & involve.aso:o & $\bullet$ &  & $\bullet$ & $\bullet$ & $\bullet$ & $\bullet$ & $\bullet$ & $\bullet$ & $\bullet$ & $\bullet$ & $\bullet$ & $\bullet$ &  & $\bullet$ & $\bullet$ & $\bullet$ & $\bullet$ & $\bullet$ & $\bullet$ &  & $\bullet$ & $\bullet$ & $\bullet$ & $\bullet$ & $\bullet$ & $\bullet$ & $\bullet$ & $\bullet$ & $\bullet$ & $\bullet$\\
0.0270 & produce.aso:o & $\bullet$ & $\bullet$ & $\bullet$ & $\bullet$ & $\bullet$ &  & $\bullet$ & $\bullet$ & $\bullet$ & $\bullet$ & $\bullet$ & $\bullet$ & $\bullet$ & $\bullet$ & $\bullet$ &  &  & $\bullet$ &  & $\bullet$ & $\bullet$ & $\bullet$ & $\bullet$ & $\bullet$ & $\bullet$ & $\bullet$ & $\bullet$ & $\bullet$ & $\bullet$ & $\bullet$\\
0.0255 & occur.as:s & $\bullet$ & $\bullet$ & $\bullet$ & $\bullet$ & $\bullet$ & $\bullet$ & $\bullet$ & $\bullet$ & $\bullet$ & $\bullet$ & $\bullet$ & $\bullet$ & $\bullet$ &  & $\bullet$ &  &  & $\bullet$ & $\bullet$ & $\bullet$ &  & $\bullet$ & $\bullet$ & $\bullet$ & $\bullet$ & $\bullet$ & $\bullet$ & $\bullet$ & $\bullet$ & \\
0.0192 & cause.aso:s & $\bullet$ & $\bullet$ & $\bullet$ & $\bullet$ & $\bullet$ & $\bullet$ & $\bullet$ &  & $\bullet$ &  & $\bullet$ & $\bullet$ & $\bullet$ &  &  & $\bullet$ &  & $\bullet$ & $\bullet$ & $\bullet$ & $\bullet$ & $\bullet$ & $\bullet$ & $\bullet$ & $\bullet$ & $\bullet$ & $\bullet$ & $\bullet$ & $\bullet$ & $\bullet$\\
0.0189 & cause.aso:o & $\bullet$ & $\bullet$ & $\bullet$ & $\bullet$ & $\bullet$ & $\bullet$ & $\bullet$ & $\bullet$ & $\bullet$ &  &  &  & $\bullet$ &  & $\bullet$ &  &  & $\bullet$ &  & $\bullet$ &  &  & $\bullet$ &  & $\bullet$ & $\bullet$ & $\bullet$ & $\bullet$ & $\bullet$ & $\bullet$\\
0.0179 & affect.aso:s & $\bullet$ & $\bullet$ & $\bullet$ & $\bullet$ & $\bullet$ & $\bullet$ & $\bullet$ & $\bullet$ & $\bullet$ & $\bullet$ & $\bullet$ & $\bullet$ & $\bullet$ &  &  & $\bullet$ & $\bullet$ & $\bullet$ & $\bullet$ & $\bullet$ & $\bullet$ & $\bullet$ & $\bullet$ & $\bullet$ & $\bullet$ & $\bullet$ & $\bullet$ & $\bullet$ & $\bullet$ & $\bullet$\\
0.0162 & require.aso:s & $\bullet$ & $\bullet$ & $\bullet$ & $\bullet$ &  & $\bullet$ & $\bullet$ & $\bullet$ & $\bullet$ & $\bullet$ & $\bullet$ & $\bullet$ & $\bullet$ & $\bullet$ &  & $\bullet$ & $\bullet$ & $\bullet$ & $\bullet$ & $\bullet$ & $\bullet$ & $\bullet$ & $\bullet$ & $\bullet$ & $\bullet$ & $\bullet$ & $\bullet$ & $\bullet$ & $\bullet$ & $\bullet$\\
0.0150 & mean.aso:o & $\bullet$ & $\bullet$ & $\bullet$ & $\bullet$ & $\bullet$ & $\bullet$ &  & $\bullet$ &  & $\bullet$ & $\bullet$ & $\bullet$ & $\bullet$ & $\bullet$ & $\bullet$ & $\bullet$ &  & $\bullet$ & $\bullet$ & $\bullet$ & $\bullet$ &  & $\bullet$ & $\bullet$ &  & $\bullet$ & $\bullet$ & $\bullet$ & $\bullet$ & $\bullet$\\
0.0140 & suggest.aso:o & $\bullet$ & $\bullet$ & $\bullet$ & $\bullet$ & $\bullet$ & $\bullet$ & $\bullet$ & $\bullet$ & $\bullet$ & $\bullet$ & $\bullet$ & $\bullet$ & $\bullet$ &  & $\bullet$ & $\bullet$ &  & $\bullet$ & $\bullet$ & $\bullet$ & $\bullet$ & $\bullet$ & $\bullet$ & $\bullet$ &  & $\bullet$ & $\bullet$ & $\bullet$ &  & $\bullet$\\
0.0138 & produce.aso:s & $\bullet$ & $\bullet$ & $\bullet$ & $\bullet$ & $\bullet$ & $\bullet$ & $\bullet$ & $\bullet$ & $\bullet$ & $\bullet$ & $\bullet$ & $\bullet$ & $\bullet$ &  &  & $\bullet$ &  & $\bullet$ & $\bullet$ & $\bullet$ &  & $\bullet$ & $\bullet$ & $\bullet$ & $\bullet$ & $\bullet$ & $\bullet$ & $\bullet$ & $\bullet$ & $\bullet$\\
0.0109 & demand.aso:o & $\bullet$ & $\bullet$ & $\bullet$ & $\bullet$ &  &  & $\bullet$ & $\bullet$ & $\bullet$ & $\bullet$ &  & $\bullet$ &  & $\bullet$ & $\bullet$ & $\bullet$ & $\bullet$ & $\bullet$ &  & $\bullet$ & $\bullet$ &  & $\bullet$ &  &  & $\bullet$ &  &  &  & $\bullet$\\
0.0109 & reduce.aso:s & $\bullet$ & $\bullet$ & $\bullet$ & $\bullet$ & $\bullet$ &  &  & $\bullet$ & $\bullet$ & $\bullet$ & $\bullet$ & $\bullet$ &  &  &  & $\bullet$ & $\bullet$ & $\bullet$ & $\bullet$ & $\bullet$ & $\bullet$ & $\bullet$ & $\bullet$ &  &  &  &  & $\bullet$ & $\bullet$ & $\bullet$\\
0.0097 & reflect.aso:o & $\bullet$ & $\bullet$ & $\bullet$ & $\bullet$ & $\bullet$ & $\bullet$ &  & $\bullet$ & $\bullet$ & $\bullet$ & $\bullet$ & $\bullet$ & $\bullet$ & $\bullet$ & $\bullet$ &  & $\bullet$ &  & $\bullet$ & $\bullet$ & $\bullet$ & $\bullet$ & $\bullet$ & $\bullet$ & $\bullet$ & $\bullet$ & $\bullet$ & $\bullet$ & $\bullet$ & $\bullet$\\
0.0092 & involve.aso:s & $\bullet$ & $\bullet$ &  &  & $\bullet$ & $\bullet$ & $\bullet$ &  & $\bullet$ & $\bullet$ & $\bullet$ & $\bullet$ &  & $\bullet$ & $\bullet$ & $\bullet$ & $\bullet$ & $\bullet$ & $\bullet$ & $\bullet$ & $\bullet$ & $\bullet$ & $\bullet$ & $\bullet$ & $\bullet$ & $\bullet$ & $\bullet$ & $\bullet$ &  & $\bullet$\\
0.0091 & undergo.aso:o & $\bullet$ & $\bullet$ & $\bullet$ & $\bullet$
 &  & $\bullet$ &  & $\bullet$ &  &  &  &  & $\bullet$ &  & $\bullet$
 & $\bullet$ &  &  & $\bullet$ &  &  &  &  &  & $\bullet$ & $\bullet$
 &  &  & $\bullet$ & $\bullet$\\
\hline\end{tabular}}

%%% Local Variables: 
%%% mode: latex
%%% TeX-master: t
%%% End: 

 \end{center}
  \caption{Class 8: dispositions}
  \label{class8}
\end{figure*}

\subsection{Lexicon Induction Experiment}

Experiments used a model with 35 classes. From maximal probability
parses for the British National Corpus derived with a statistical
parser \cite{CarrollRooth:98}, we extracted frequency tables for intransitive
verb/subject pairs and transitive verb/subject/object triples. The 500
most frequent verbs were selected for slot labeling. Fig.
\ref{ILexEx1} shows two verbs $v$ for which the most probable class
label is 5, a class which we earlier described as \emph{communicative
  action}, together with the estimated frequencies
of $f(n)p_\theta(c|n)$ for those ten nouns $n$ for which this estimated
frequency is highest. 

\begin{figure}[tb]
  \begin{center}
    {\small
      \begin{tabular}{|ll||ll|}
        \hline
        \emph{blush} 5 & 0.982975 & \emph{snarl} 5  & 0.962094 \\
        \hline
        constance & 3 & mandeville & 2 \\
        christina & 3 & jinkwa & 2 \\
        willie & 2.99737 & man & 1.99859\\
        ronni & 2 & scott & 1.99761\\
        claudia & 2 & omalley & 1.99755 \\
        gabriel & 2 & shamlou & 1 \\
        maggie & 2 & angalo & 1\\
        bathsheba & 2 & corbett & 1\\
        sarah & 2 & southgate & 1\\
        girl & 1.9977 & ace & 1\\
        \hline
      \end{tabular}
      }
  \end{center}
  \caption{Lexicon entries: \emph{blush}, \emph{snarl}}
  \label{ILexEx1}
\end{figure}

Fig. \ref{ILexEx2} shows corresponding data for an intransitive scalar
motion sense of {\em increase}.

\begin{figure}[tb]
  \begin{center}    
    {\small
      \begin{tabular}{|ll|ll|}
        \hline
       \emph{increase} 17 & 0.923698 & & \\
      \hline
      number & 134.147 &       proportion & 23.8699 \\
      demand & 30.7322 &      size & 22.8108 \\
      pressure & 30.5844 &      rate & 20.9593 \\
      temperature & 25.9691 &      level & 20.7651 \\
      cost & 23.9431 &      price & 17.9996 \\
      \hline
    \end{tabular}
    }
\end{center}
\caption{Scalar motion \emph{increase}.}
\label{ILexEx2}
\end{figure}

Fig. \ref{scalar-change-verbs} shows the intransitive verbs which take
17 as the most probable label.  Intuitively, the verbs are
semantically coherent. When compared to \newcite{Levin:93}'s 48
top-level verb classes, we found an agreement of our classification
with her class of ``verbs of changes of state'' except for the last
three verbs in the list in Fig. \ref{scalar-change-verbs} which is
sorted by probability of the class label.

Similar results for German intransitive scalar motion verbs are shown
in Fig. \ref{steigerungsverben}. The data for these experiments were
extracted from the maximal-probability parses of a 4.1 million word
corpus of German subordinate clauses, yielding 418290 tokens (318086
types) of pairs of verbs or adjectives and nouns. The lexicalized
probabilistic grammar for German used is described in
\newcite{Beil:99}. We compared the German example of scalar motion
verbs to the linguistic classification of verbs given by
\newcite{Schuhmacher:86} and found an agreement of our classification
with the class of ``einfache \"Anderungsverben'' (simple verbs of
change) except for the verbs \emph{anwachsen} (increase) and
\emph{stagnieren}(stagnate) which were not classified there at all.

\begin{figure}[tb]
  \begin{center}
    {\small
      \begin{tabular}{|ll|ll|}
        \hline
        0.977992 & decrease & 0.560727 & drop \\
        0.948099 & double & 0.476524 & grow \\
        0.923698 & increase & 0.42842 & vary \\
        0.908378 & decline & 0.365586 & improve \\
        0.877338 & rise & 0.365374 & climb \\
        0.876083 & soar & 0.292716 & flow \\
        0.803479 & fall  & 0.280183 & cut \\
        0.672409 & slow & 0.238182 & mount \\
        0.583314 & diminish & & \\
        \hline
      \end{tabular}
      }
  \end{center}
  \caption{Scalar motion verbs}
  \label{scalar-change-verbs}
\end{figure}

\begin{figure}[tb]
  \begin{center}
{\small
    \begin{tabular}{|ll|l|l|}
      \hline
      0.741467   &      ansteigen& (go up)\\
      0.720221   &      steigen& (rise)\\
      0.693922   &      absinken& (sink)\\
      0.656021   &      sinken& (go down)\\
      0.438486   &      schrumpfen& (shrink)\\
      0.375039   &      zurückgehen& (decrease)\\
      0.316081    &     anwachsen& (increase)\\
      0.215156    &     stagnieren& (stagnate)\\
      0.160317    &     wachsen& (grow)\\
      0.154633     &    hinzukommen & (be added) \\
      \hline
    \end{tabular}
}
  \end{center}
  \caption{German intransitive scalar motion verbs}
  \label{steigerungsverben}
\end{figure}

Fig. \ref{TLexEx1} shows the most probable pair of classes for {\em
increase} as a transitive verb, together with estimated frequencies
for the head filler pair.  Note that the object label 17 is the
class found with intransitive scalar motion verbs; this correspondence
is exploited in the next section.

\begin{figure}[tb]
  \begin{center}
    {\small
      \begin{tabular}{|ll|}
        \hline
        \emph{increase} $(8,17)$ &    0.3097650 \\   \hline
        development -  pressure &  2.3055  \\
        fat -  risk  &     2.11807  \\
        communication -  awareness   &     2.04227  \\
        supplementation -  concentration &  1.98918  \\
        increase -  number    &    1.80559  \\
        \hline
      \end{tabular}
      }
  \end{center}
  \caption{Transitive \emph{increase} with estimated frequencies for
  filler pairs.}
  \label{TLexEx1}
\end{figure}

\section{Linguistic Interpretation}

\begin{figure}
\begin{center}
\mbox{\psfig{file=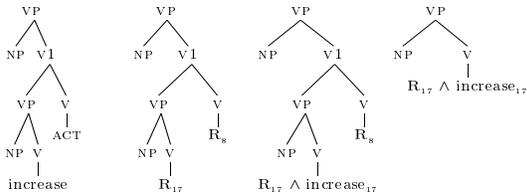,width=7cm,height=2.5cm}}
\caption{First tree: linguistic lexical entry for transitive
 verb {\em increase}. Second: corresponding lexical entry
with induced classes as relational constants.  Third:
indexed open class root added as conjunct in
transitive scalar motion {\em increase}.
Fourth: induced entry
for related intransitive {\em increase}.
    }
\label{lextrees}
\end{center}
\end{figure}

\newcommand{\sub}[1]{\raisebox{-0.6ex}{{\tiny{#1}}}} 
In some linguistic accounts, multi-place verbs are decomposed into
representations involving (at least) one predicate or relation per
argument.  For instance, the transitive causative/inchoative verb {\em
  increase,} is composed of an actor/causative verb combining
with a one-place predicate in the structure on the left in
Fig.~\ref{lextrees}. Linguistically, such representations are motivated by
argument alternations (diathesis), case linking and deep word order,
language acquistion, scope ambiguity, by the desire to represent
aspects of lexical meaning, and by the fact that in some languages,
the postulated decomposed representations are overt, with each
primitive predicate corresponding to a morpheme.  For references and
recent discussion of this kind of theory see \newcite{HaleKeyser:93}
and \newcite{Kural:96}.

We will sketch an understanding of the lexical representations induced
by latent-class labeling in terms of the linguistic theories mentioned
above, aiming at an interpretation which combines computational
learnability, linguistic motivation, and denotational-semantic
adequacy.  The basic idea is that latent classes are computational
models of the atomic relation symbols occurring in lexical-semantic
representations.  As a first implementation, consider replacing
the relation symbols in the first tree in Fig.~\ref{lextrees} with relation
symbols derived from the latent class labeling. In the second tree in
Fig \ref{lextrees}, $R_{17}$ and $R_{8}$ are relation symbols with indices
derived from the labeling procedure of Sect. \ref{s:hsc}. Such
representations can be semantically interpreted in standard ways, for
instance by interpreting relation symbols as denoting relations
between events and individuals.

Such representations are semantically inadequate for reasons given in
philosophical critiques of decomposed linguistic representations; see
\newcite{Fodor:98} for recent discussion.  A lexicon estimated in the above
way has as many primitive relations as there are latent classes.  We
guess there should be a few hundred classes in an approximately
complete lexicon (which would have to be estimated from a corpus of
hundreds of millions of words or more). Fodor's arguments, which are
based on the very limited degree of genuine interdefinability of
lexical items and on Putnam's arguments for contextual
determination of lexical meaning, indicate that the number of basic
concepts has the order of magnitude of the lexicon itself. More
concretely, a lexicon constructed along the above principles would
identify verbs which are labelled with the same latent classes; for
instance it might identify the representations of {\em grab} and {\em
  touch}.

For these reasons, a semantically adequate lexicon must include
additional relational constants. We meet this requirement in a simple
way, by including as a conjunct a unique constant derived from the
open-class root, as in the third tree in Fig. \ref{lextrees}.  We
introduce indexing of the open class root (copied from the class
index) in order that homophony of open class roots not result in
common conjuncts in semantic representations---for instance, we don't
want the two senses of {\em decline} exemplified in {\em decline the
proposal} and {\em decline five percent} to have a common entailment
represented by a common conjunct. This indexing method works as long
as the labeling process produces different latent class labels for the
different senses.

The last tree in Fig. \ref{lextrees} is the learned representation for
the scalar motion sense of the intransitive verb {\em increase}. In
our approach, learning the argument alternation (diathesis) relating
the transitive {\em increase} (in its scalar motion sense) to the
intransitive {\em increase} (in its scalar motion sense) amounts to
learning representations with a common component $R_{17} \wedge 
{\rm increase\sub{17}}$.  In this case, this is achieved.

\section{Conclusion}

We have proposed a procedure which maps observations of
subcategorization frames with their complement fillers to structured
lexical entries.  We believe the method is
scientifically interesting, practically useful, and
flexible because: 
\begin{enumerate}
\item  The algorithms and implementation are efficient enough to map a corpus of a
  hundred million words to a lexicon.
\item The model and induction algorithm have foundations in the
  theory of parameterized families of probability distributions and statistical
  estimation.  As exemplified in
  the paper, learning, disambiguation, and evaluation
  can be given simple, motivated formulations.
\item The derived lexical representations are linguistically
  interpretable.  This suggests the possibility of large-scale
  modeling and observational experiments bearing on questions arising
  in linguistic theories of the lexicon.
\item Because a simple probabilistic model is used, the induced
  lexical entries could be incorporated in lexicalized syntax-based
  probabilistic language models, in particular in head-lexicalized
  models.  This provides for potential application in many areas.
\item The method is applicable to any natural language where text
  samples of sufficient size, computational morphology,
  and a robust parser capable of extracting
  subcategorization frames with their fillers are available. 
\end{enumerate}

\end{document}